\documentclass[conference]{IEEEtran}
\IEEEoverridecommandlockouts
\usepackage{cite}
\usepackage{amsmath,amssymb,amsfonts}
\usepackage{algorithmic}
\usepackage{graphicx}
\usepackage{textcomp}
\usepackage{xcolor}
\usepackage{graphicx}
\usepackage{multirow}
\usepackage{url}
\usepackage{threeparttable}
\usepackage{fancyhdr}
\usepackage{amsmath,amsfonts}
\usepackage{tikz}
\usetikzlibrary{shapes}
\usetikzlibrary{calc}
\usepackage{stackengine,scalerel}
\def\BibTeX{{\rm B\kern-.05em{\sc i\kern-.025em b}\kern-.08em
    T\kern-.1667em\lower.7ex\hbox{E}\kern-.125emX}}

\begin{document}

\title{Is Smaller Always Faster? Tradeoffs in Compressing Self-Supervised Speech Transformers}

\author{
Tzu-Quan Lin$^1$, 
Tsung-Huan Yang$^2$$^*$,
Chun-Yao Chang$^3$$^*$,
Kuang-Ming Chen$^4$$^*$,\thanks{$^*$This work was done while the authors were at National Taiwan University.}\\
Tzu-hsun Feng$^1$,
Hung-yi Lee$^1$, 
Hao Tang$^5$\\
$^1$Graduate Institute of Communication Engineering, National Taiwan University, Taiwan\\
$^2$Academia Sinica, Taiwan\\
$^3$University of California, Los Angeles, United States\\
$^4$University of Washington, United States\\
$^5$University of Edinburgh, United Kingdom
}

\maketitle

\begin{abstract}
Transformer-based self-supervised models have achieved remarkable success in speech processing, but their large size and high inference cost present significant challenges for real-world deployment. While numerous compression techniques have been proposed, inconsistent evaluation metrics make it difficult to compare their practical effectiveness. In this work, we conduct a comprehensive study of four common compression methods, including weight pruning, head pruning, low-rank approximation, and knowledge distillation on self-supervised speech Transformers. We evaluate each method under three key metrics: parameter count, multiply-accumulate operations, and real-time factor. Results show that each method offers distinct advantages. In addition, we contextualize recent compression techniques, comparing DistilHuBERT, FitHuBERT, LightHuBERT, ARMHuBERT, and STaRHuBERT under the same framework, offering practical guidance on compression for deployment. 
\end{abstract}


\vspace{-0.5em}
\section{Introduction}

Self-supervised learning (SSL) has driven significant progress in speech processing, enabling models to learn powerful representations from large unlabeled datasets~\cite{schneider2019wav2vec, chung2019unsupervised, baevski2020wav2vec, hsu2021hubert}. However, these models are typically large and compute-intensive, making them difficult to deploy in real-time or on resource-constrained devices. Model compression is therefore essential to bridge the gap between research prototypes and practical deployment.

Numerous techniques have been proposed for compressing SSL models, including weight pruning~\cite{han2016deepcompress}, low-rank approximation~\cite{denton2014els}, and knowledge distillation~\cite{hinton2015distilling}. Yet, these methods are often evaluated in isolation and compared under different metrics, such as parameter count, multiply-accumulate operations (MACs), or inference time. This inconsistency makes it difficult to assess their practical strengths and tradeoffs. A method that reduces parameter count might not improve actual inference speed, and one that accelerates GPU runtime might have little impact on theoretical complexity. Without a consistent evaluation framework, it remains unclear which method is best suited for which use case.

In this paper, we conduct a comprehensive study on compressing Transformer-based speech SSL models. We systematically compare four classic compression methods—weight pruning, head pruning, low-rank approximation, and knowledge distillation—under three critical metrics: number of parameters, MACs, and real-time factor. 

Our results show that each method exhibits distinct strengths under different metrics: weight pruning significantly reduces parameter count and MACs; head pruning improves inference speed by reducing attention computations; low-rank approximation effectively reduces parameter counts; and distillation provides the most balanced tradeoff between efficiency and performance.

To further clarify the efficiency-performance tradeoffs in recent research, we also contextualize several representative compression methods, including DistilHuBERT~\cite{chang2022distilhubert}, FitHuBERT~\cite{lee2022fithubert}, LightHuBERT~\cite{wang2022lighthubert}, ARMHuBERT~\cite{jang2023recycle}, and STaRHuBERT~\cite{jang2023star} within the same framework. This unified comparison not only highlights how design decisions impact compression outcomes, but also offers practical guidance for selecting methods based on deployment constraints.

\section{Related Work}

Model compression has been widely studied in the context of NLP and computer vision, particularly for Transformer-based architectures. Common strategies include knowledge distillation~\cite{hinton2015distilling, sanh2019distilbert}, weight pruning~\cite{gordon2020compressing}, head pruning~\cite{michel2019sixteen}, low-rank approximation~\cite{denton2014els}, and early exiting~\cite{fan2019reducing, liu2020fastbert}.
In the speech domain, early efforts on compressing SSL models have largely focused on task-specific settings, where compression is jointly optimized with downstream objectives~\cite{lai2021parp, peng2021shrinking, fu2023distillw2v2, peng2023structured}. 

Recent studies have proposed task-agnostic compression methods that aim to preserve the general usability of speech representations.
DistilHuBERT~\cite{chang2022distilhubert} distills a 12-layer HuBERT into a shallow 2-layer student by aligning hidden representations using cosine and $\ell_1$ losses. 
FitHuBERT~\cite{lee2022fithubert} retains the model depth but reduces the width of attention heads, feedforward layers, and convolutional channels to achieve a thinner architecture. 
LightHuBERT~\cite{wang2022lighthubert} introduces a two-stage training process: first distilling a large model into a supernet, then applying once-for-all training to support a range of subnetwork configurations.
ARMHuBERT~\cite{jang2023recycle} and STaRHuBERT~\cite{jang2023star} share a similar architecture with FitHuBERT but achieve superior performance through masking distillation and temporal relation distillation, respectively.
Beyond distillation-based approaches, several task-agnostic, pruning-based methods have been proposed for speech self-supervised models~\cite{wang2023task, lin2024property, lin2025identifying}.
Additionally, Lin et al.~\cite{lin2024daisy} propose a method that dynamically determines whether to perform early exiting during feature extraction.

While these prior works report promising results, they each adopt different evaluation metrics—DistilHuBERT, LightHuBERT, and ARMHuBERT report only parameter count; FitHuBERT reports parameter count and real-time factor; and STaRHuBERT reports parameter count and MACs—making it difficult to directly compare their efficiency-performance tradeoffs. Our unified framework addresses this gap by providing consistent and comprehensive evaluations.

\section{Model Compression}

We plan to study four classic compression methods, weight pruning, head pruning, low-rank approximation, and knowledge distillation.
All except distillation are instances of iterative pruning, where pruning and retraining alternate until convergence.
We intentionally exclude more involved compression strategies that integrate task objectives or joint pruning components in order to isolate and clarify the effect of each technique.

After pruning, model performance typically degrades due to increased training loss. To recover performance, additional training is applied. To track recovery, we monitor the self-supervised loss curve before and after compression. We adopt MelHuBERT~\cite{lin2022melhubert} as our base model for its simple training recipe, which allows efficient pre-training from scratch and loss monitoring.

\subsection{MelHuBERT}
MelHuBERT simplifies HuBERT by using Mel spectrograms as input and cross-entropy as the training loss. It removes waveform-level convolutions and accelerates training. 
MelHuBERT saves about 33.3\% of MACs
and about 31.2\% of pre-training time on a single 24GB RTX 3090 GPU.
Given a Mel-spectrogram sequence $x_1 \dots x_T$, we quantize each frame into cluster indices $c_t$ using k-means. The training objective is:
\begin{align}
\mathcal{L} = -\sum_{t=1}^T \log p(c_t | x),
\end{align}
\begin{align}
     \quad p(c_t|x) = \frac{\exp(W_{c_t}^\top o_t)}{\sum_{i=1}^K \exp(W_i^\top o_t)},\label{eq:melhubert}
\end{align}
where $W$ is the classifier matrix and $o_t$ the model output. This loss is used throughout our compression analysis.

\subsection{Weight Pruning}
Weight pruning is one of the earliest model compression techniques~\cite{lecun1990obd}, based on the observation that small weights can be removed with minimal impact on performance, and retraining can recover any loss. In iterative pruning, we progressively remove low-magnitude weights and fine-tune the model. We adopt a simple magnitude-based criterion~\cite{han2015learning, frankle2021lth}, where weights are masked as follows:
\begin{align}
    m_i = \begin{cases}
        0 & \text{if } |\theta_i| < \delta \\
        1 & \text{otherwise}
    \end{cases}
\end{align}
where $\delta$ is a threshold set by the target sparsity. The forward pass becomes $f(x; m \odot \theta)$, where $m$ is the binary mask. While weight pruning significantly reduces parameters and MACs, the unstructured sparsity limits real-time gains on standard hardware.

\subsection{Head Pruning}
Head pruning targets Transformer-specific structure by removing full attention heads, each comprising its own query, key, and value matrices. Prior work~\cite{voita2019analyzing, michel2019sixteen} shows that many heads are redundant. We evaluate the impact of pruning heads in speech SSL models. While several methods have been proposed, we adopt a gradient-based importance score from~\cite{michel2019sixteen}, since simpler $\ell_1$-norm-based scores perform consistently worse in our experiments. 
Specifically, we use a small data set $\{(x_1, y_1), \dots, (x_n, y_n)\}$
to compute the score
\begin{align}
s_i = \sum_{k=1}^n \left\| \text{vec}\left(J_i(x_k)^\top \frac{\partial \mathcal{L}}{\partial J_i}(J_i(x_k), y_k)\right) \right\|_1
\end{align}
for each head $H_i$,
where $J_i$ is the attention map multiplied by the forward result of the value matrix $V_i$,
and $\frac{\partial \mathcal{L}}{\partial J_i}$ is the gradient to that particular computation node.

\subsection{Low-Rank Approximation}
Low-rank approximation leverages the empirical low-rank nature of learned weight matrices~\cite{tara2013lrmf}. Rather than decomposing projection matrices, we prune intermediate dimensions in feed-forward layers, where the hidden size is typically large (e.g., 3072). Let $U \in \mathbb{R}^{d \times 3072}$ and $V \in \mathbb{R}^{3072 \times d}$ be the input/output projections. We compute a score for each hidden unit $i$ as:
\begin{align}
    s_i = \sum_{k=1}^{d} (|U_{ki}| + |V_{ik}|).
\end{align}
We iteratively remove units with the lowest scores. Unlike conventional low-rank factorization, this direct pruning approach simplifies implementation while achieving comparable efficiency.

\subsection{Knowledge Distillation}
Knowledge distillation compresses models by training a small student to imitate a large teacher~\cite{ba2014deep}. It allows full architectural flexibility. We distill from a 12-layer MelHuBERT into 2- and 6-layer students using KL divergence on output distributions:
\begin{align}
    \frac{1}{T} \sum_{t=1}^T \text{KL}[p(c_t|x) \| \hat{p}(c_t|x)]
\end{align}
where $p(c_t | x)$ is the distribution predicted by the teacher defined in \eqref{eq:melhubert},
\begin{align}
    \hat{p}(c_t|x) = \frac{\exp(\hat{W}_{c_t}^\top o_t)}{\sum_{i=1}^K \exp(\hat{W}_i^\top o_t)}
\end{align}
is the distribution of centroids predicted by the student,
and $\hat{W}$ is a trainable projection matrix of the student.
All the parameters of the teacher are frozen during distillation.

Unlike pruning-based methods, distillation trains a new model from scratch and does not preserve teacher weights or structure. It is widely used in prior SSL compression works~\cite{chang2022distilhubert, lee2022fithubert, wang2022lighthubert}.

\section{Design of Experiments}

To examine compression tradeoffs, we pre-train and compress 12-layer Transformers on the full 960-hour LibriSpeech dataset. Representations are evaluated on phoneme recognition (LibriSpeech 100-hour subset) and speaker identification (VoxCeleb1), following SUPERB~\cite{shu2021superb} with frozen encoders and weighted-sum layer aggregation. Pre-training loss serves as a proxy for downstream performance in pruning experiments.

\vspace{0.5em}
\subsection{MelHuBERT}
We adopt MelHuBERT-10ms and -20ms~\cite{lin2022melhubert}, pre-trained in two stages with masked prediction~\cite{hsu2021hubert, chen2022wavlm}. Stage 1 uses k-means (512 clusters) on log Mel features to train a 12-layer Transformer for frame-wise prediction (10ms) or every-other-frame prediction (20ms). Stage 2 applies k-means on the 9th-layer features from Stage 1, with frame-wise prediction in both variants.

Masking follows~\cite{lin2022melhubert}: 7\% (10 frames) for 10ms and 14\% (5 frames) for 20ms. Stage 1 is trained for 200 epochs (both variants); Stage 2 is trained for 200 epochs (10ms) and 100 epochs (20ms), using 8 V100 GPUs, Adam optimizer (lr=$10^{-4}$), batch size 32, and 10\% dropout after key/value/query matrices and FC layers.

\vspace{0.5em}
\subsection{Weight Pruning}
For weight pruning, we iteratively prune individual weights
based on their $\ell_1$ for all weights and biases in linear layers of Transformers.
We keep track of the exponential moving average of the loss with a decay of 0.9998,
and if the loss does not change much (within 0.001) compared to the one 15,000 steps before,
pruning is triggered.
We use a pruning schedule that is aggressive when the network is dense,
and mild when the network is sparse.
In particular, we prune 20\% until 80\% density,
10\% until 50\% density, 5\% until 35\% density, 2.5\% until 30\% density,
1\% until 10\% density, and 0.5\% until 5\% density.
We use a batch size of 4 and a learning rate of $10^{-5}$.
The pruning schedule follows~\cite{frankle2021lth,chen2021lottery}
with minor modifications.

\vspace{0.5em}
\subsection{Head Pruning}
For head pruning, we use 25\% of the training data to
compute the scores of each head.
Similarly, since the $\ell_1$ norm varies across layers,
as opposed to pruning a fixed number from each layer,
we normalize the scores of heads within each layer
and prune the heads by comparing them all together.
As for all iterative pruning approaches, we train the model
for a fixed 25,000 steps after pruning,
with learning rate $10^{-5}$ and a batch size of 4.

\begin{figure*}
  \begin{center}
  \definecolor{pr}{HTML}{1F77B4}
  \definecolor{sid}{HTML}{FF7F0E}
  \definecolor{loss}{HTML}{2CA02C}
  \begin{tikzpicture}[font=\footnotesize]
  \draw[loss, thick] (0, 0) -- (0.8, 0);
  \draw[loss] (0.33, 0.07) -- (0.47, -0.07);
  \draw[loss] (0.33, -0.07) -- (0.47, 0.07);
  \node[right] at (0.8, 0) {Pre-training loss};

  \draw[pr, thick] (3.0, 0) -- (3.8, 0);
  \fill[pr] (3.4, 0) circle (2pt);
  \node[right] at (3.8, 0) {Phone recognition};
  
  \draw[sid, thick] (6.2, 0) -- (7.0, 0);
  \node[fill=sid, regular polygon, regular polygon sides=3, inner sep=1pt] at (6.6, 0) {};
  \node[right] at (7.0, 0) {Speaker identification};
  \end{tikzpicture}

  \begin{tikzpicture} \node[rotate=90] at (0, 0) {(a) 10 ms}; \node at (0, -1.5) {}; \end{tikzpicture}
  \includegraphics[height=3.3cm]{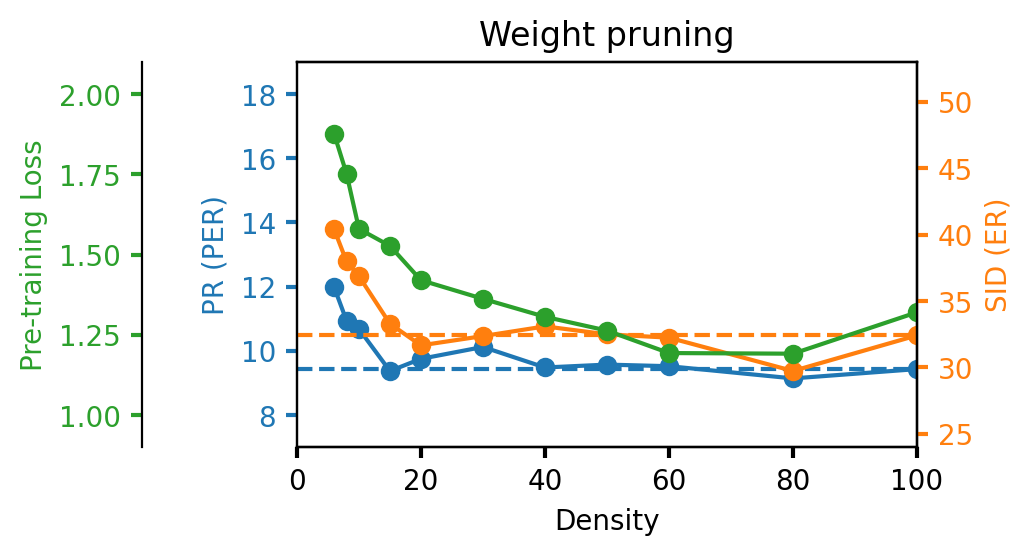}
  \includegraphics[height=3.3cm]{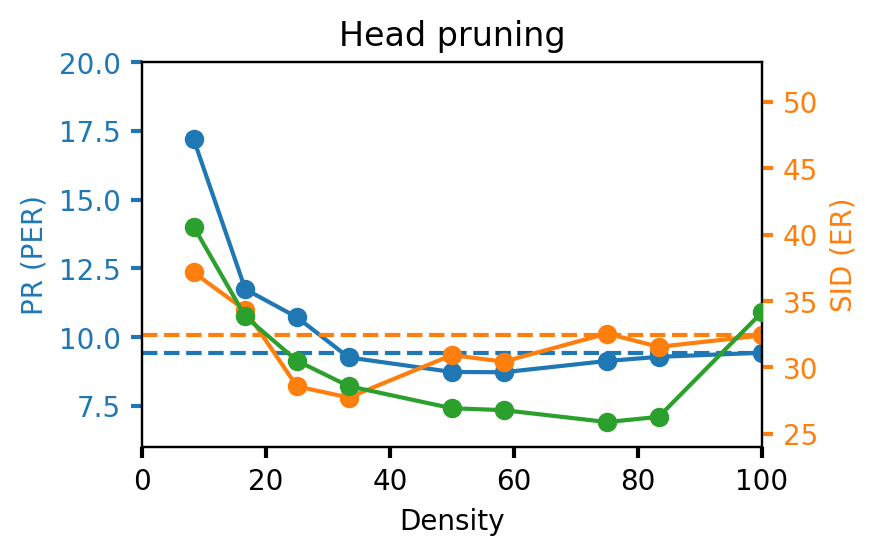}
  \includegraphics[height=3.3cm]{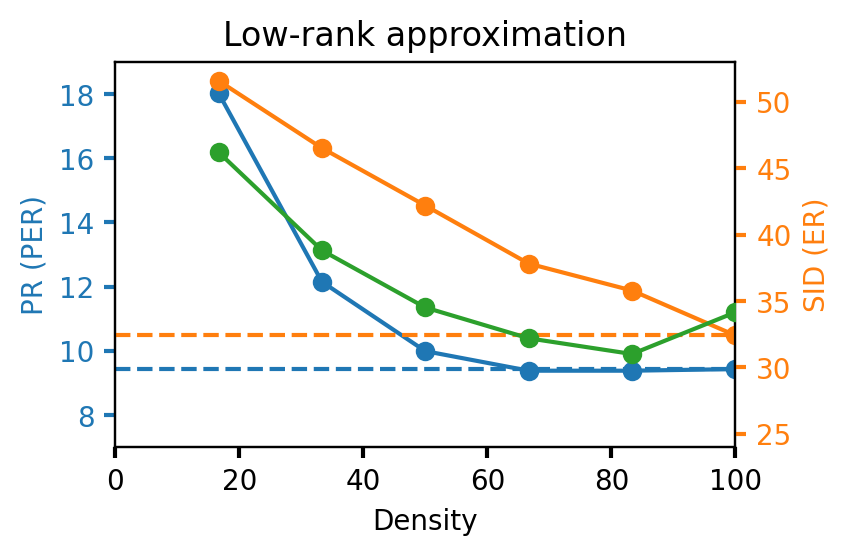} \\
  \begin{tikzpicture} \node[rotate=90] at (0, 0) {(b) 20 ms}; \node at (0, -1.5) {}; \end{tikzpicture}
  \includegraphics[height=3.3cm]{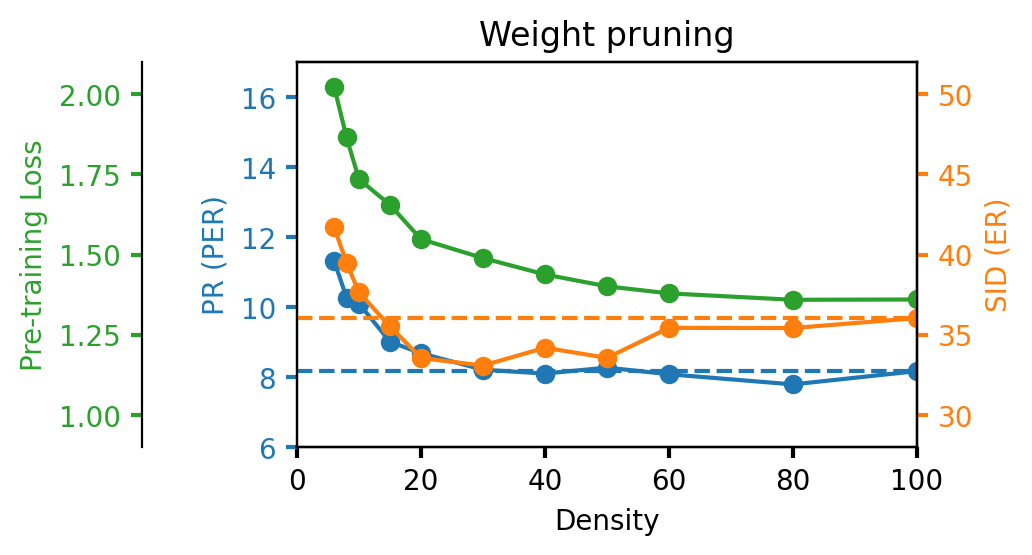}
  \includegraphics[height=3.3cm]{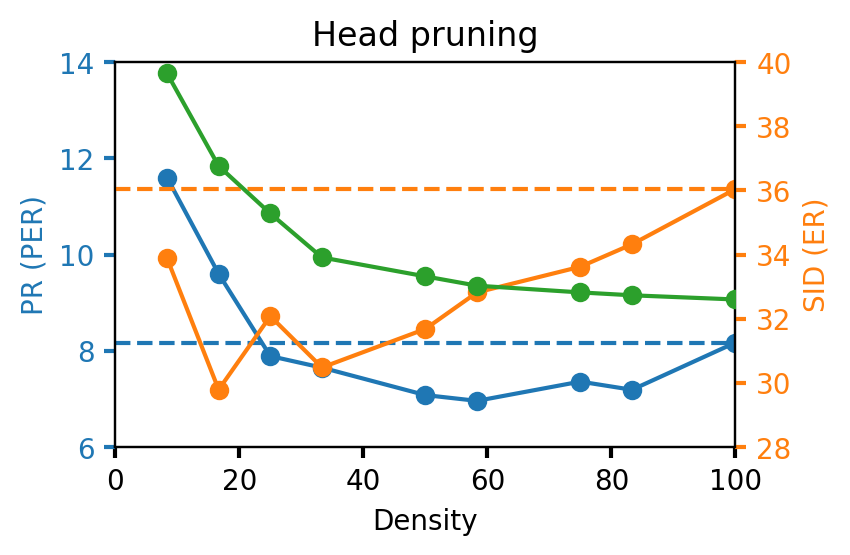}
  \includegraphics[height=3.3cm]{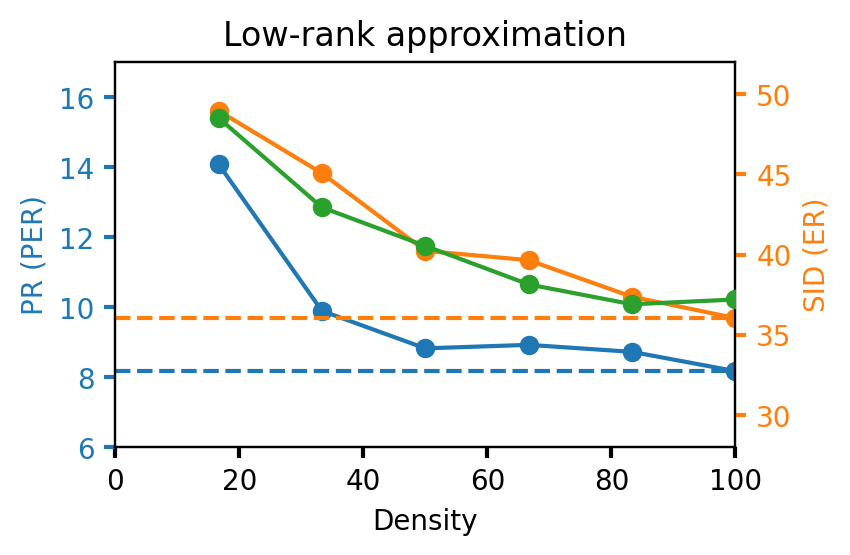}
  \vspace{-0.5em}
  \caption{The downstream performance of compressed MelHuBERT-10ms
    and MelHuBERT-20ms on PR and SID for weight pruning, head pruning,
    and low-rank approximation. The dashed lines show the performance of the
    unpruned MelHuBERT. The densities are relative to each pruning
    technique. For weight pruning, the density is relative to
    the remaining weights. For head pruning, the density is relative to
    the total number of heads. For low-rank approximation, the density is
    relative to the hidden dimensions of the feedforward layers.
    \label{fig:downstream}}
  \end{center}
\end{figure*}

\begin{figure}
  \centering
  \begin{tabular}{cc}
  \includegraphics[width=4.2cm]{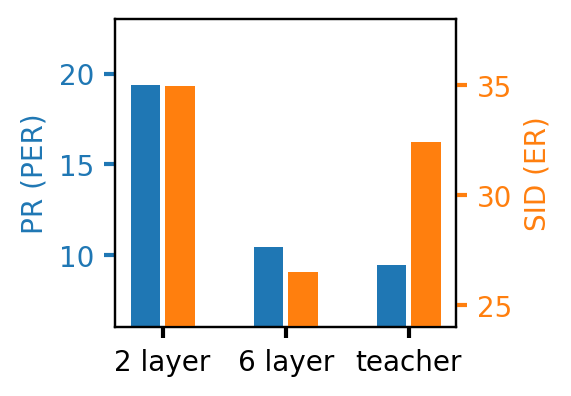} &
  \includegraphics[width=4.2cm]{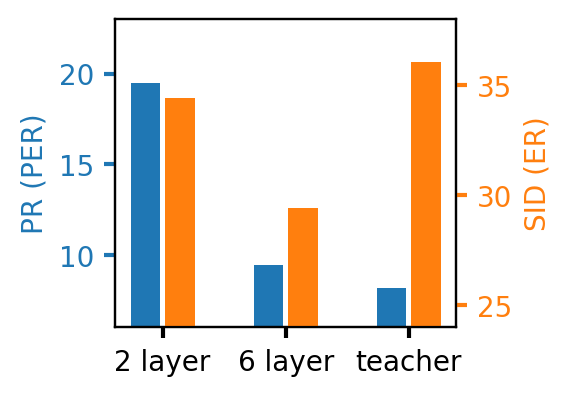} \\
  (a) 10 ms &
  (b) 20 ms
  \end{tabular}
  \caption{The downstream performance of 2-layer and 6-layer Transformers
    distilling from MelHuBERT-10ms and MelHuBERT-20ms.
    \label{fig:kd}}
\end{figure}

\vspace{0.5em}
\subsection{Low-Rank Approximation}
For low-rank approximation, recall that we target the 3072-
dimensional output of the FC1 layer. Reducing the dimension
amounts to pruning the rows of FC1 and the columns of
FC2 based on the sum of their $\ell_1$ norm. The input and
output dimensions (in our case 768) are seldom larger than
the dimension in the middle (in this case 3072).
We prune 128 dimensions
every 25,000 steps, and use a learning rate of $10^{-5}$ and a
batch size of 4 to train the model after every pruning step.

\vspace{0.5em}
\subsection{Knowledge Distillation}
For knowledge distillation, we explore a 2-layer and a 6-layer student
network.  
Following prior works~\cite{wang2022lighthubert, jang2023recycle}, we study several design decisions,
including changing the masking strategy, initializing with MelHuBERT layers, and the temperature when training with KL divergence.  
Based on our study, neither applying masking on the input nor initializing with
MelHuBERT layers leads to improvement.  
Hence, we simply minimize KL
divergence as our distillation loss without masking and initializing.

\section{Results}

Given the design of individual compression approaches,
in this section, we present a comprehensive comparison across
approaches.
Since each individual approach comes with its own strengths
and weaknesses, it is likely difficult to conclude definitively
which approach is the best under all conditions.
Instead, we evaluate them with several metrics to showcase
their strengths and weaknesses.

\begin{figure*}
  \definecolor{melhubert}{HTML}{FF7F0E}
  \definecolor{know-dist}{HTML}{2CA02C}
  \definecolor{head-pruning}{HTML}{D62728}
  \definecolor{row-pruning}{HTML}{9467BD}
  \definecolor{weight-pruning}{HTML}{8C564B}
  \begin{center}
  \begin{tikzpicture}[font=\footnotesize]
  \draw[melhubert, thick] (0, 0) -- (0.8, 0);
  \filldraw[melhubert] (0.4, 0) circle (2pt);
  \node[right, align=left] at (0.8, 0) (node1) {MelHuBERT \\ MelHuBERT-first2 \\ MelHuBERT-first6};
  
  \draw[weight-pruning, thick] (3.4, 0) -- (4.2, 0);
  \node[fill, weight-pruning, regular polygon, regular polygon sides=3, rotate=90, inner sep=1.2pt] at (3.8, 0) {};
  \node[right] at (4.2, 0) (node5) {weight pruning};

  \draw[head-pruning,thick] (6.2, 0) -- (7.0, 0);
  \node[fill, head-pruning, regular polygon, regular polygon sides=3, inner sep=1.2pt] at (6.6, 0) {};
  \node[right] at (7.0, 0) (node3) {head pruning};
  
  \draw[row-pruning, thick] (8.8, 0) -- (9.6, 0);
  \node[fill, row-pruning, regular polygon, regular polygon sides=3, rotate=180, inner sep=1.2pt] at (9.2, 0) {};
  \node[right, align=left] at (9.6, 0) (node4) {low-rank \\ approximation};
  
  \draw[know-dist, thick] (11.5, 0) -- (12.3, 0);
  \draw[know-dist] (11.82, 0.08) -- (11.98, -0.08);
  \draw[know-dist] (11.82, -0.08) -- (11.98, 0.08);
  \node[right] at (12.3, 0) (node2) {knowledge distillation};
  
  \end{tikzpicture}
  \end{center}
  \begin{tikzpicture} \node[rotate=90] at (0, 0) {(a) 10 ms}; \node at (0, -1.5) {}; \end{tikzpicture}
  \includegraphics[width=5.5cm]{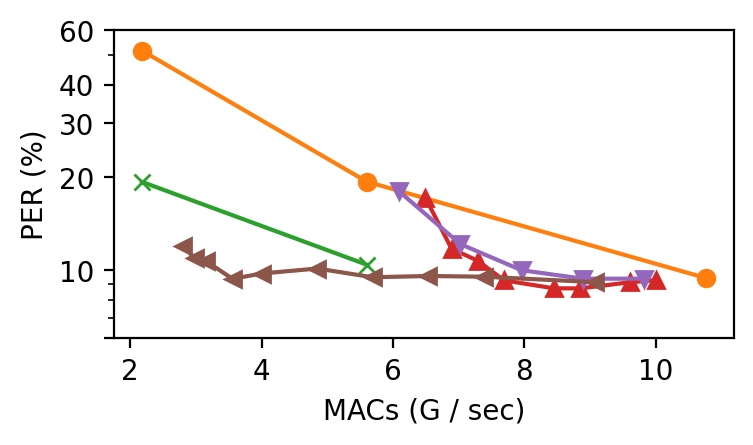}
  \includegraphics[width=5.5cm]{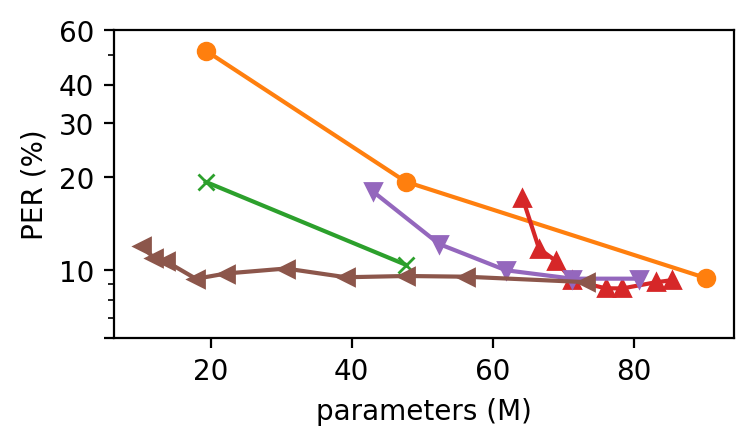}
  \includegraphics[width=5.5cm]{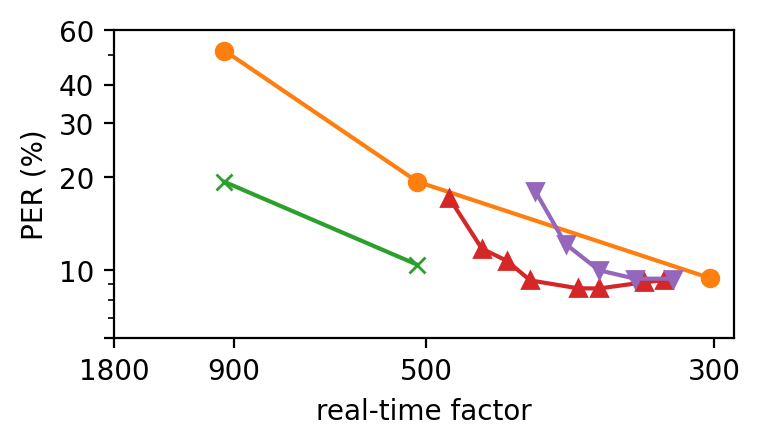} \\
  \begin{tikzpicture} \node[rotate=90] at (0, 0) {(b) 20 ms}; \node at (0, -1.5) {}; \end{tikzpicture}
  \includegraphics[width=5.5cm]{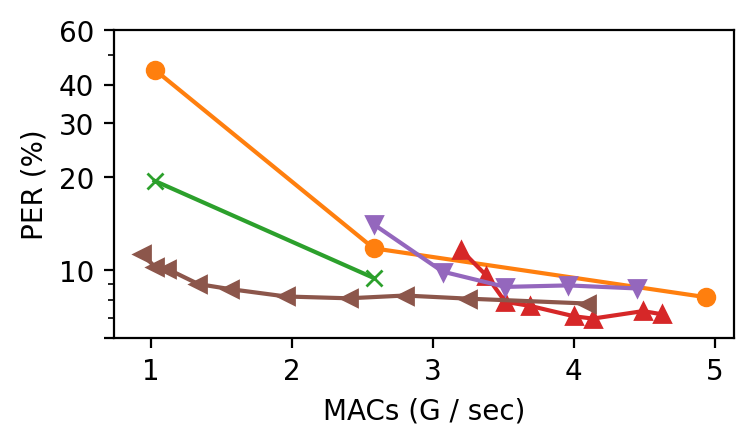}
  \includegraphics[width=5.5cm]{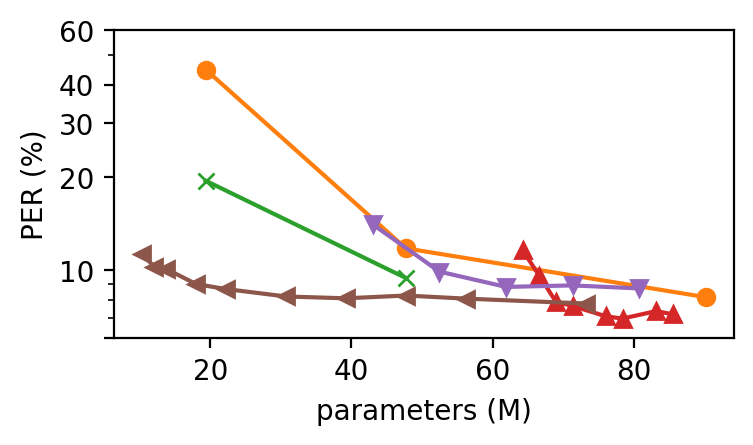}
  \includegraphics[width=5.5cm]{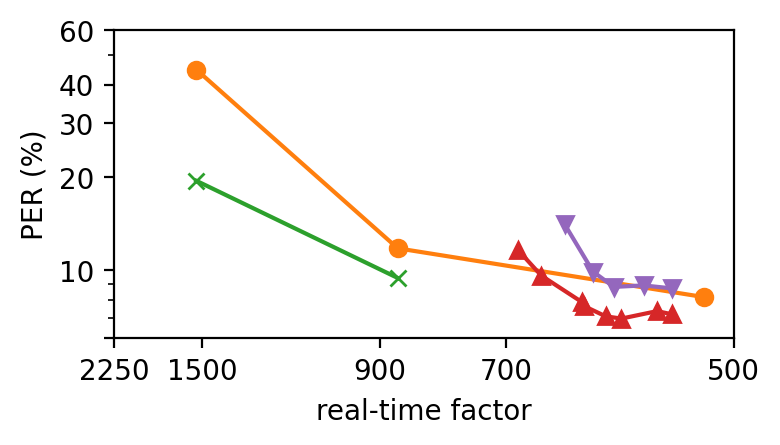}
  \vspace{-0.5em}
  \caption{A comparison of four compression methods,
    namely weight pruning, head pruning, low-rank approximation, and knowledge
    distillation, when applied to MelHuBERT-10ms (top row) and MelHuBERT-20ms (bottom row).
    We also compare with stopping the forward process at the second (MelHuBERT-first2)
    and the sixth (MelHuBERT-first6) layer of MelHuBERT.
    The metrics include MACs per one second speech (first column), numbers of parameters (second column),
    and real-time factor (third column).
    \label{fig:macs-params-runtime}}
\end{figure*}

\vspace{0.5em}
\subsection{Downstream performance}

We first study downstream performance, specifically
phone recognition and speaker identification, along the pruning process,
i.e., at different densities.
It is noted that density is measured relative to the total amount that
can be pruned.
For example, in head pruning, density is the number of attention
heads left over the total number of heads in the unpruned Transformer.
We are interested in to what extent downstream performance is maintained
while we prune models with the pre-training loss.
The results of the iterative pruning family are presented in Figure~\ref{fig:downstream}, while those of knowledge distillation are shown in Figure~\ref{fig:kd}.

With weight pruning, we can prune the model to at least 40\% density
while maintaining phone recognition performance under both 10ms and 20ms
frame rates.
Similarly, we can prune the model to 20\% density while
maintaining speaker identification performance under both frame rates.
As we have shown in the last section, as pruning proceeds, there is a
period where the pre-training loss and the downstream performance both
improve over the unpruned model.

With head pruning, we can prune the model to at least 33\% density while
maintaining the phone recognition and speaker identification performance
under both 10ms and 20ms frame rates.
Similar to weight pruning, as pruning proceeds,
there is a period where the downstream performance is
better than the unpruned model.

For low-rank approximation however, the results are different from weight
pruning and head pruning---downstream performance starts to degrade, as
soon as pruning starts.  For phone recognition, we can prune the rows
and columns in the feed-forward layers to at least 50\% density while
keeping the degradation within 0.5\% PER.  However, the impact of low-rank
approximation on speaker identification is much stronger than on phone
recognition.
We only focus on pruning with $\ell_1$ norm, so there might exist
better pruning approaches for the feed-forward layers.
Regardless, this suggests that the feed-forward layers in Transformers
are particularly important for speaker identification.

For knowledge distillation, we find that both the 2-layer and
6-layer students are able to retain the performance of teachers on speaker
identification, with the 6-layer ones performing even better than the
teachers.  
Neither student is able to match the teachers’ performance on phone recognition, with the 2-layer one lagging significantly behind.
The worse performance seems to suggest that knowledge distillation
might not be an effective compression approach.  However, as we will
see in later sections, once we put all the results in context, knowledge
distillation actually performs favorably compared to others.

\vspace{0.5em}
\subsection{MACs, parameters, and runtime}

While Figure \ref{fig:downstream} and Figure \ref{fig:kd} show how well each approach maintains
performance after compression, they do not show the benefit of
compression.
Density is a relative measure, making it difficult to compare
across approaches.
To evaluate the benefit of compression across approaches,
we decide to use three metrics, the number of multiply-accumulation operations (MACs)
per one second speech, the number of parameters, and
the runtime measured on a single GPU.
The three metrics are related.
Reduction in MACs represents the theoretical speed-up of
inference when parallel computation is not available.
In the presence of a GPU, the reduction in MACs does not always
translate to faster runtime, as it depends on
the structure of the pruned networks.
Reduction in model parameters usually leads to a reduction in MACs
and can improve generalization.
As we have seen in weight pruning and head pruning (Figure~\ref{fig:downstream}),
there is a period along the pruning process where
the pre-training loss and downstream performance 
both improve.
Results of all approaches evaluated on the three metrics
are shown in Figure~\ref{fig:macs-params-runtime}.
We only focus on phone recognition, but it shows a similar
trend for speaker identification.

First, we find that weight pruning has a significant effect on
reducing MACs per one second speech and the number of parameters,
without affecting phone recognition much.
However, the resulting networks are sparse, and it is difficult to
parallelize computation unless dedicated hardware is designed
for the pruned network.
As a result, we do not measure runtime on the GPU for weight
pruning.
For serial computation, weight pruning is the best choice
among all approaches.

In Transformers, the pairs of feed-forward layers constitute
a significant number of parameters, so
the reduction in the number of parameters achieved through
head pruning is relatively limited compared to the other three approaches.
However, head pruning provides a significant runtime speed-up on GPU,
due to the expensive computation involved in self-attention.
The improvement is particularly pronounced in the 10 ms case,
as the input in the 10 ms case is twice as long as the 20 ms one.

\begin{figure*}
  \definecolor{melhubert}{HTML}{FFBF86}
  \definecolor{know-dist}{HTML}{95CF95}
  \definecolor{head-pruning}{HTML}{EA9393}
  \definecolor{row-pruning}{HTML}{C9B3DE}
  \definecolor{weight-pruning}{HTML}{C5AAA5}
  \definecolor{combined}{HTML}{E377C2}
  \begin{center}
  \begin{tikzpicture}[font=\footnotesize]
  \draw[melhubert, thick] (0, 0) -- (0.8, 0);
  \filldraw[melhubert] (0.4, 0) circle (2pt);
  \node[right, align=left] at (0.8, 0) (node1) {MelHuBERT \\ MelHuBERT-first2 \\ MelHuBERT-first6};
  
  \draw[weight-pruning, thick] (3.4, 0) -- (4.2, 0);
  \node[fill, weight-pruning, regular polygon, regular polygon sides=3, rotate=90, inner sep=1.2pt] at (3.8, 0) {};
  \node[right] at (4.2, 0) (node5) {weight pruning};

  \draw[head-pruning,thick] (6.2, 0) -- (7.0, 0);
  \node[fill, head-pruning, regular polygon, regular polygon sides=3, inner sep=1.2pt] at (6.6, 0) {};
  \node[right] at (7.0, 0) (node3) {head pruning};
  
  \draw[row-pruning, thick] (8.8, 0) -- (9.6, 0);
  \node[fill, row-pruning, regular polygon, regular polygon sides=3, rotate=180, inner sep=1.2pt] at (9.2, 0) {};
  \node[right, align=left] at (9.6, 0) (node4) {low-rank \\ approximation};
  
  \draw[know-dist, thick] (11.5, 0) -- (12.3, 0);
  \draw[know-dist] (11.82, 0.08) -- (11.98, -0.08);
  \draw[know-dist] (11.82, -0.08) -- (11.98, 0.08);
  \node[right] at (12.3, 0) (node2) {knowledge distillation};

  \draw[combined, thick] (0, -1) -- (0.8, -1);
  \node[fill=combined, regular polygon, regular polygon sides=4, inner sep=1.4pt] at (0.4, -1) {};
  \node[right] at (0.8, -1) {knowledge distillation followed by head pruning and low-rank approximation};
  
  \end{tikzpicture}

  \includegraphics[width=5.5cm]{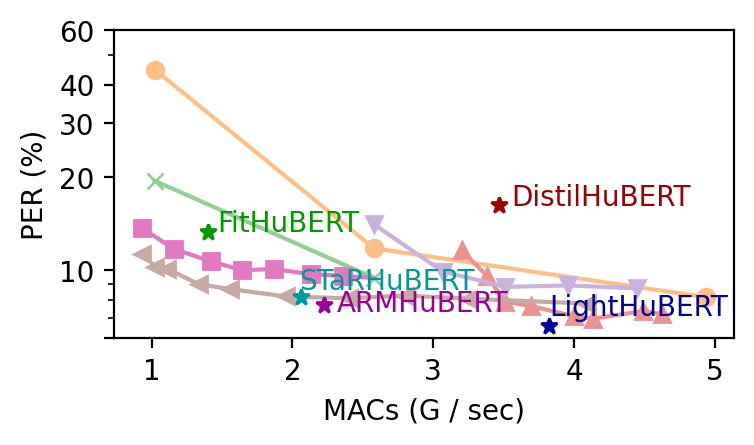}
  \includegraphics[width=5.5cm]{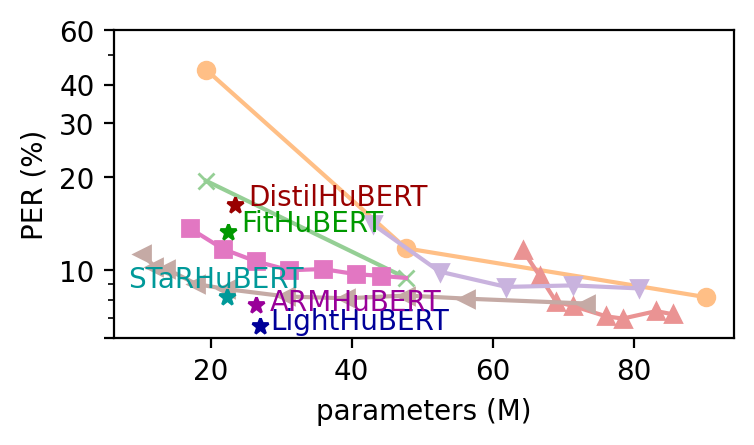}
  \includegraphics[width=5.5cm]{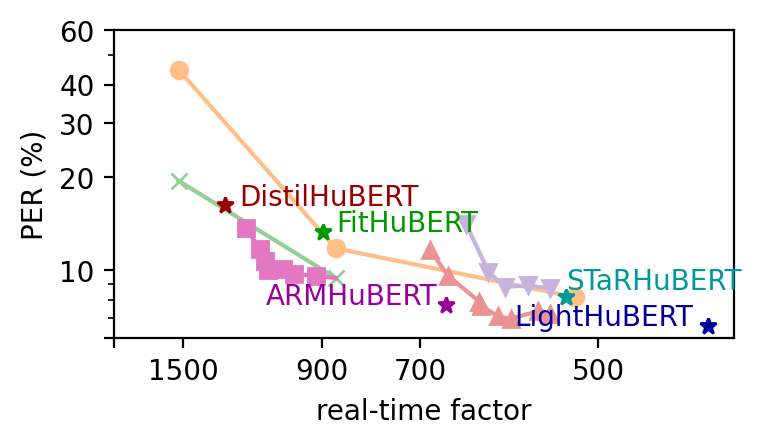}

  \end{center}
  
  \caption{Comparing DistilHuBERT, FitHuBERT, and LightHuBERT to compressing MelHuBERT-20m
    in terms of MACs per one second speech, numbers of parameters, and real-time factors
    on phone recognition.
    \label{fig:comparison}}
\end{figure*}

Because the parameters are concentrated in the pairs of feed-forward layers,
low-rank approximation provides a favorable reduction in
the number of parameters.
However, the feed-forward layers are significantly faster to compute
on a GPU than self-attention.
Even though low-rank approximation might achieve a lower MACs per
one second speech compared to head pruning, the reduction in runtime
is limited.

Knowledge distillation (with a 2-layer and a 6-layer student)
achieves a significant reduction on all three metrics.
In fact, knowledge distillation provides the best tradeoff
among all approaches.
We attribute this to the number of GPU kernel calls.
Each GPU kernel call has a significant overhead, and
computing with fewer layers not only reduces MACs per one second
speech, but also saves the overhead.
For example, a model pruned with low-rank approximation
can have a similar MACs per one second speech
compared to a 6-layer student,
but the runtime of the pruned model is significantly slower
than the 6-layer student.

Finally, inspired by the results of knowledge distillation, we provide a baseline that only takes the first two layers
and the first six layers of the pre-trained Transformers.
This amounts to stopping the forward process at the second
and the sixth layer.
This approach involves no additional training and is
arguably the simplest baseline.
Combined with other approaches in Figure~\ref{fig:macs-params-runtime},
we now have a better picture of what the worst and the best are able to achieve.

\vspace{0.5em}
\subsection{Contextualizing prior work}

In addition to the four classic compression methods adopted in this study, several more advanced techniques have been proposed for compressing self-supervised Transformers. However, most of these methods emphasize maintaining performance across a wide range of tasks rather than focusing on the tradeoffs discussed in the previous section. To address this gap, we selected five HuBERT-based compression methods for comparison: DistilHuBERT~\cite{chang2022distilhubert}, FitHuBERT~\cite{lee2022fithubert}, LightHuBERT~\cite{wang2022lighthubert}, ARMHuBERT~\cite{jang2023recycle}, and STaRHuBERT~\cite{jang2023star}. Our findings allow us not only to identify their respective strengths and weaknesses but also to explain the reasons behind them.

Figure \ref{fig:comparison} illustrates the performance of these five methods compared to the classic compression approaches we study, evaluated on phone recognition. Among them, LightHuBERT achieves the lowest phone error rate (PER). This is largely because LightHuBERT undergoes an initial round of self-distillation before compression, resulting in a strong baseline.

Despite this performance, both DistilHuBERT and LightHuBERT consume significantly more MACs per second of speech due to retaining the convolutional layers from the original HuBERT model. Specifically, in LightHuBERT, convolutional layers account for approximately 64\% of MACs while contributing only 4.4\% of the model's total parameters. In contrast, the convolutional layers of FitHuBERT, ARMHuBERT, and STaRHuBERT are optimized, leading to significantly lower MACs per second.

While these five methods have a similar number of parameters, their phone recognition performance varies significantly. The difference is primarily attributed to model depth—DistilHuBERT has only two layers, whereas FitHuBERT, LightHuBERT, ARMHuBERT, and STaRHuBERT have twelve. This result indicates that for phone recognition, the depth of Transformers is more critical than the number of heads or the size of the feed-forward layers.

However, in terms of actual runtime, LightHuBERT is the slowest, while FitHuBERT offers no significant improvement over simply using the first six layers of MelHuBERT. As discussed in the previous section, increasing the number of layers results in more GPU kernel calls, where the overhead of these calls can become a bottleneck, especially when each call is relatively lightweight. Consequently, DistilHuBERT achieves a notable runtime reduction by limiting the model to two layers.

ARMHuBERT demonstrates a clear runtime advantage over STaRHuBERT, despite their similar MACs and parameter counts. This advantage arises from ARMHuBERT’s ability to reuse attention maps for every two layers, significantly reducing the computation cost of self-attention—a result that aligns with our analysis in the previous section.

In summary, our analysis contextualizes the performance of DistilHuBERT, FitHuBERT, LightHuBERT, ARMHuBERT, and STaRHuBERT, providing insights into their respective strengths and weaknesses.

\vspace{0.5em}
\subsection{Combining compression techniques}

Given our findings, we can combine compression techniques
to make use of the strengths of each technique.
We first use knowledge distillation to learn a
6-layer student from a 12-layer MelHuBERT-20ms.
Six layers strike the right balance between
reducing the overhead of GPU kernel calls
and downstream performance.
We then perform two iterations of head pruning, reducing the number of heads from 72 to 36.
We further perform five iterations of low-rank approximation, reducing the dimension from 3072 to 512. 

Both head pruning and low-rank approximation
are effective at reducing MACs per one second speech,
numbers of parameters, and real-time factors.
All hyperparameters remain the same as introduced
in early sections.

The results of combining compression techniques
are shown in Figure~\ref{fig:comparison}.
The combined approach is strictly better than
knowledge distillation alone,
and is close to weight pruning on MACs per
one second speech and numbers of parameters.
In addition, the resulting models are also
strictly better than DistilHuBERT and FitHuBERT.
Compared to dedicated approaches, simple combination
of compression techniques is able to produce
strong results.

\section{Conclusion}

Based on the weight pruning results, we can conclude that there exist small but strong self-supervised models that can produce representations that generalize across tasks.
Weight pruning, even the vanilla version used in this work, is in fact able to find models much smaller than the recent approaches.
However, smaller models do not necessarily run faster.
Even though runtime is correlated with MACs, the characteristics of the hardware (such as the amount of parallelization) play a much more important role.
In other words, if the goal is to improve runtime, reducing the number of GPU kernel calls is much more effective than reducing the size of the kernel calls.
We hope that this study encourages future work to consider multiple dimensions when designing and evaluating compression strategies.

\bibliographystyle{IEEEtran}
\bibliography{mybib}

\end{document}